\documentclass[final]{cvpr}
\usepackage{times}
\usepackage{graphicx}
\usepackage{amsmath}
\usepackage{amssymb}
\usepackage{booktabs}
\usepackage{tabularx}

\usepackage[pagebackref=true,breaklinks=true,colorlinks,bookmarks=false]{hyperref}

\def\cvprPaperID{4554} % *** Enter the CVPR Paper ID here
\def\confYear{CVPR 2021}

\newcolumntype{Y}{>{\centering\arraybackslash}X}

\begin{document}

%%%%%%%%% TITLE
\title{Single-View 3D Object Reconstruction from Shape Priors in Memory}

\author{Shuo~Yang\textsuperscript{\rm 1}\qquad~Min~Xu\textsuperscript{\rm 1}\thanks{Corresponding Author.}\qquad~Haozhe~Xie\textsuperscript{\rm 2}\qquad~Stuart~Perry\textsuperscript{\rm 1}\qquad~Jiahao~Xia\textsuperscript{\rm 1}\qquad\\
		{\textsuperscript{\rm 1} School of Electrical and Data Engineering, University of Technology Sydney}  \\
		{\textsuperscript{\rm 2} Harbin Institute of Technology}\\
		\small{\texttt{\{shuo.yang,jiahao.xia\}@student.uts.edu.au}} \qquad\\
		\small{\texttt{\{min.xu,stuart.perry\}@uts.edu.au}} \qquad
		\small{\texttt{cshzxie@gmail.com}}
	}

\maketitle

\begin{abstract}
Existing methods for single-view 3D object reconstruction directly learn to transform image features into 3D representations. However, these methods are vulnerable to images containing noisy backgrounds and heavy occlusions because the extracted image features do not contain enough information to reconstruct high-quality 3D shapes. Humans routinely use incomplete or noisy visual cues from an image to retrieve similar 3D shapes from their memory and reconstruct the 3D shape of an object. Inspired by this, we propose a novel method, named Mem3D, that explicitly constructs shape priors to supplement the missing information in the image. Specifically, the shape priors are in the forms of ``image-voxel'' pairs in the memory network, which is stored by a well-designed writing strategy during training. We also propose a voxel triplet loss function that helps to retrieve the precise 3D shapes that are highly related to the input image from shape priors. The LSTM-based shape encoder is introduced to extract information from the retrieved 3D shapes, which are useful in recovering the 3D shape of an object that is heavily occluded or in complex environments. Experimental results demonstrate that Mem3D significantly improves reconstruction quality and performs favorably against state-of-the-art methods on the ShapeNet and Pix3D datasets. 
\end{abstract}
\section{Introduction}

\begin{figure}[!ht]
  \resizebox{\linewidth}{!} {
    \includegraphics{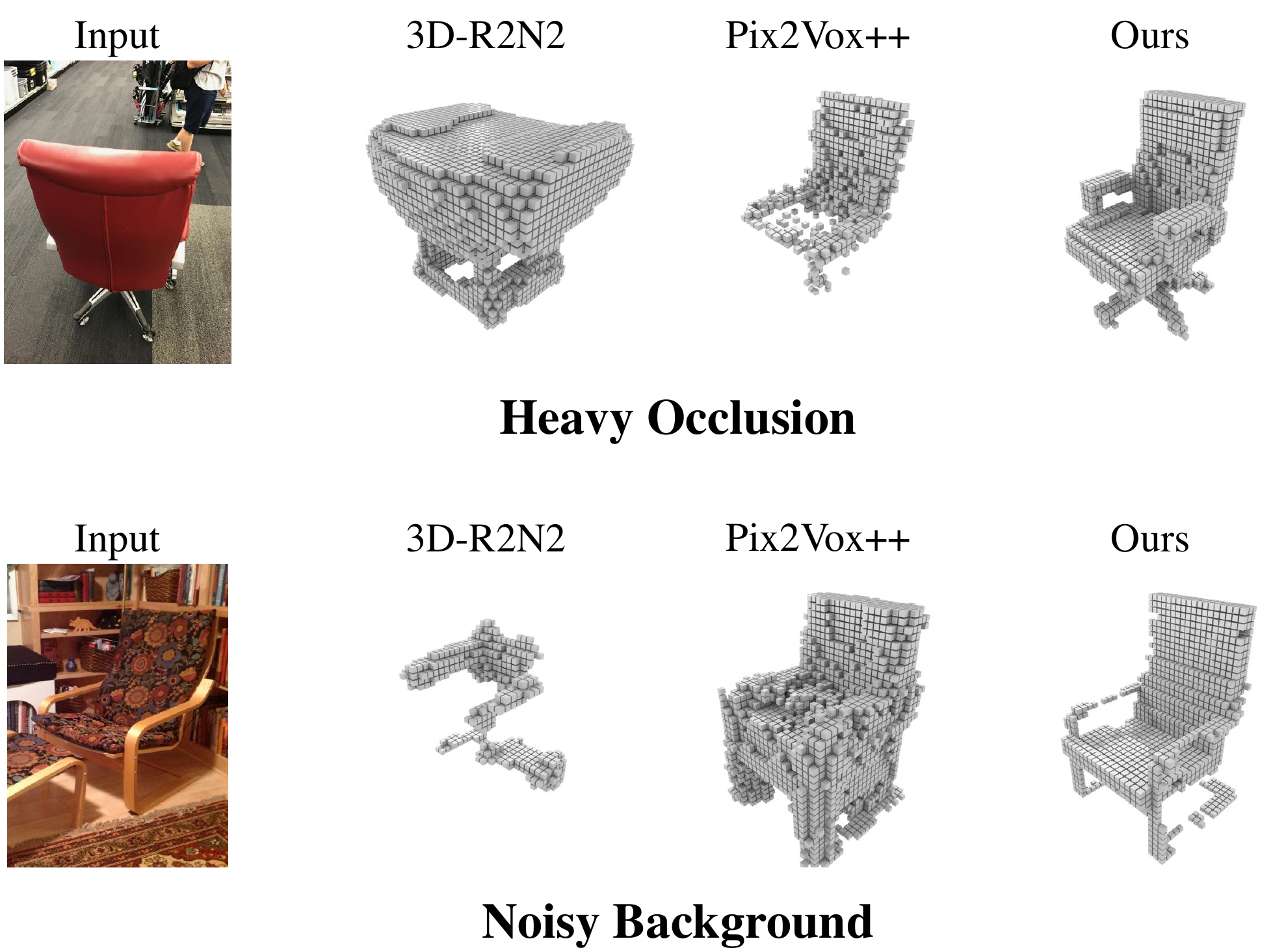}
  }
  \caption{Compared with the current state-of-the-arts method Pix2Vox++~\cite{Pix2Vox++} and classic method 3D-R2N2~\cite{DBLP:conf/eccv/ChoyXGCS16}, the proposed method are more robust in reconstructing the 3D shape of an object from a single image that contains occlusion or noisy backgrounds.}
    \label{fig:issues}
\end{figure}

% Reconstructing object 3D shape from a single-view RGB image is a vital but challenging computer vision task in robotics, CAD, and virtual and augmented reality.
% Humans can easily infer the 3D shape of an object from a single image due to sufficient prior knowledge and ability of visual understanding and reasoning, while it is an extremely difficult and ill-posed problem for a machine vision system because a single-view image can not provide sufficient information for the object to be reconstructed.
Reconstructing object 3D shape from a single-view RGB image is a vital but challenging computer vision task in
robotics, CAD, and virtual and augmented reality applications. 
Humans can easily infer the 3D shape of an object from a single image due to sufficient prior knowledge and an innate ability for visual understanding and reasoning. 
However, this is an extremely difficult and ill-posed problem for a machine vision systems because a single-view image can not provide sufficient information for the object to be reconstructed.

Most of the existing learning-based methods for single-view 3D reconstruction extract features from a single RGB image, then transform it into a 3D representation.
These methods achieve promising results on the synthetic datasets (ShapeNet~\cite{DBLP:journals/corr/ChangFGHHLSSSSX15}).
However, as shown in Figure~\ref{fig:issues}, they usually have trouble reconstructing the 3D shape of an object from real-world images.
The performance gap between the real-world and synthetic datasets are caused by the quality of image features.
The features extracted from images with noisy backgrounds or heavy occlusions usually contain insufficient useful information for 3D reconstruction.

Humans can infer a reasonable 3D shape of an object from a single image, even with incomplete or noisy visual cues.
This is due to the fact that humans retrieve similar shapes from their memories and apply these shape priors to recover the shape of hidden and noisy parts of the object.
Motivated by human vision, we propose a novel memory-based framework for 3D reconstruction, Mem3D, which consists of four components: image encoder, memory network, LSTM shape encoder, and shape decoder.
The proposed Mem3D explicitly constructs the shape priors in the memory network that help complete the missing image features to recover the 3D shape of an object that is heavy occluded or in a complex environment.
To construct the shape priors, we design a writing strategy to store the ``image-voxel'' pairs into the memory network in a key-value fashion during training.
To retrieve the precise 3D shapes that are highly related to the input image from the memory network, we propose a voxel triplet loss function that guarantees that images with similar 3D shapes are closer in the feature space.
To better leverage the retrieved shapes, the LSTM-based shape encoder transforms the useful knowledge of these shapes into a shape prior vector.
To employ both information from image and shape priors, the input image features and the output of the LSTM-based shape encoder are concatenated and are forwarded to a decoder to predict the 3D shape of the object.

The main contributions are summarized as follows:

\begin{itemize}
  \item We propose a memory-based framework for single-view 3D object reconstruction, named Mem3D. It innovatively retrieves similar 3D shapes from the constructed shape priors, and shows a powerful ability to reconstruct the 3D shape of objects that are heavily occluded or in a complex environment.
  \item We present a memory network that stores shape priors in the form of ``image-voxel'' pairs. To better organize the shape priors and ensure accurate retrieval, we design novel reading and writing strategies, as well as introducing a voxel triplet loss function.
  \item Experimental results demonstrate that the proposed Mem3D significantly improves the reconstruction quality and performs favorably against state-of-the-art methods on the ShapeNet and Pix3D datasets. 
 \end{itemize}
 
 \section{Related Work}

\noindent \textbf{Single-image 3D Reconstruction.}
Recently, 3D object reconstruction from a single-view image has attracted increasing attention because of its wide applications in the real world.
Recovering object shape from a single-view image is an ill-posed problem due to the limitation of visual clues. Existing works use the representation of silhouettes~\cite{Dibra_2017_CVPR}, shading~\cite{Richter_2015_CVPR}, and texture~\cite{DBLP:journals/ai/Witkin81} to recover 3D shape.
With the success of deep learning, especially generative adversarial networks~\cite{DBLP:conf/nips/GoodfellowPMXWOCB14} and variational autoencoders~\cite{DBLP:journals/corr/KingmaW13}, the deep neural network based encoder-decoder has become the main-stream architecture, such as 3D-VAE-GAN~\cite{DBLP:conf/nips/0001ZXFT16}. 
PSGN~\cite{DBLP:conf/cvpr/FanSG17} and 3DLMNet~\cite{DBLP:conf/bmvc/MandikalLAR18} generate point representations from single-view images.
3D-R2N2~\cite{DBLP:conf/eccv/ChoyXGCS16} is a unified framework for single- and multi-view 3D reconstruction which employs a 3D convolutional LSTM to fuse the image features.
To solve the permutation variance issue, Pix2Vox~\cite{Xie_2019_ICCV} employs a context-aware fusion module to adaptively select high-quality reconstructions from single-view reconstructions.
However, these works that utilize shape priors implicitly are venerable to noisy backgrounds and heavy occlusions.
To reconstruct the 3D shape of an object from real-world images, 
MarrNet~\cite{DBLP:conf/nips/0001WXSFT17} and its variants~\cite{shapehd,DBLP:conf/nips/ZhangZZTF018} reconstructs 3D objects by estimating depth, surface normals, and silhouettes.
Both 3D-RCNN~\cite{DBLP:conf/cvpr/KunduLR18} and FroDO~\cite{FroDo} introduce an object detector to remove noisy backgrounds.

\noindent \textbf{Memory Network.}
The Memory Network was first proposed in~\cite{weston2014memory}, which augmented neural networks with an external memory module that enables the neural network to store long-term memory.
Later works~\cite{DBLP:conf/icml/KumarIOIBGZPS16,DBLP:conf/nips/SukhbaatarSWF15} improve the Memory Network so it can be trained in an end-to-end manner.
Hierarchical Memory Networks~\cite{ch2016hierarchical} was proposed to allow a read controller to efficiently access large scale memories.
Key-Value Memory Networks~\cite{DBLP:journals/corr/MillerFDKBW16} store prior knowledge in a key-value structured memory, where keys are used to address relevant memories whose corresponding values are returned. 

\section{Method}
 
\begin{figure*}
    \resizebox{\linewidth}{!} {
      \includegraphics{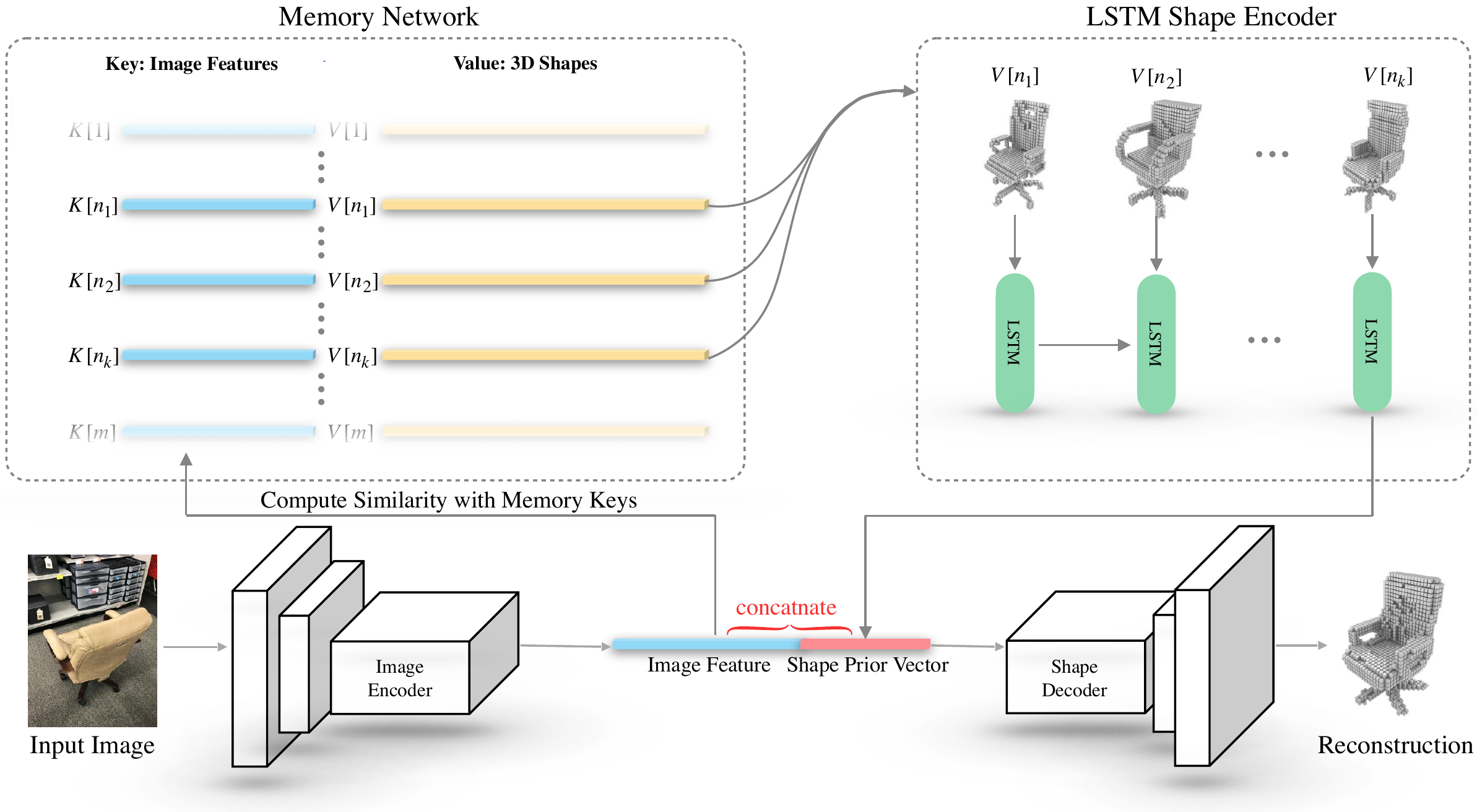}
    }
    \caption{The proposed Mem3D reconstruct the 3D shape of an object from a single input image. The Memory Network learns to retrieve 3D volumes that are highly related to the input image. The LSTM Shape Encoder is proposed to contextually encode multiple 3D volumes into a shape prior vector, which provides the information that helps to recover the 3D shape of the object's hidden and noisy parts.}
    \label{fig:flow}
\end{figure*}

In existing single-view 3D reconstruction methods~\cite{Pix2Vox++,wang2018pixel2mesh,AttSets,DBLP:conf/eccv/ChoyXGCS16}, the shape priors are learnt into model parameters, which leads to low quality reconstructions for images containing heavy occlusion and noisy backgrounds.
To alleviate this issue, the proposed Mem3D explicitly constructs the shape priors using a Key-Value Memory Network~\cite{DBLP:conf/emnlp/MillerFDKBW16}.
Specifically, the image encoder extracts features from the input image.
During training, the extracted features and the corresponding 3D shape are then stored in the memory network in a key-value fashion. 
For both training and testing, the 3D shapes whose corresponding keys have high similarities are forwarded to the LSTM shape encoder.
After that, the LSTM shape encoder generates a shape prior vector.
Finally, the decoder takes the both image features and the shape prior vector to reconstruct the 3D shape of the object. 

\subsection{Memory Network}
\label{memory}

The memory network aims to explicitly construct the shape priors by storing the ``image-voxel'' pairs, which memorize the correspondence between the image features and the corresponding 3D shapes.The memory items are constructed as: [key, value, age], which is denoted as $M = \left\{(\mathbf{K}_i, \mathbf{V}_i, A_i)_{i=1}^m\right\}$,  where $m$ denotes the size of the memory.
The ``key'' and ``value'' memory slots store the image features and the corresponding 3D volume, respectively.
The ``key'' $\mathbf{K}_i \in \mathbb{R}^{n_k}$ is used to compute the cosine similarities with the input image features.
The ``value'' $\mathbf{V}_i \in \mathbb{R}^{n_v}$ is returned if the similarity score between the query and the keys of memory exceeds a threshold. The $n_k$ and $n_v$ are dimension of the memory ``key'' and memory ``value'', respectively.
The ``age'' $\mathbf{A}_i \in \mathbb{N}$ represents the alive time of the pair, which is to set to zero when the pair is matched by the input image features.
The memory network overwrites the ``oldest'' pair when writing new pairs.
\subsubsection{Memory Writer}

The memory writer is presented to construct the shape priors in the memory network.
We designed a writing strategy to determine how to update the memory slots when given the image features $\mathbf{F} \in \mathbb{R}^{n_k}$ and its corresponding volume $\mathcal{V}$. 
The memory writing only works at training because it takes the ground truth 3D volumes as input, which are not available during testing.
In the memory network, the key similarity between the input image features $\mathbf{F}$ and the memory key $\mathbf{K}_i$ is defined as following

\begin{equation}\label{eq:key_sim}
  S_k(\mathbf{F}, \mathbf{K}_i) = { \frac{\mathbf{F} \cdot \mathbf{K}_{i}}{\lVert  \mathbf{F} \lVert  \lVert \mathbf{K}_{i} \lVert} }
\end{equation}

Similarly, the value similarity between the corresponding 3D volumes $\mathcal{V}$ and the value $\mathbf{V}_i$ can be defined as

\begin{equation}\label{eq:value_sim}
  S_v(\mathcal{V}, \mathbf{V}_i) =  1 - \frac{1}{r_v^3}\sum_{j=1}^{r_v^3}(\mathbf{V}_i^j-\mathcal{V}^j)^2
\end{equation}
where $r_v$ indicates the resolution of the 3D volume.

The writing strategy works for the two cases according to whether the similarity satisfies $S_v(\mathcal{V}, \mathbf{V}_{n_1}) > \delta$, where $\delta$ is the similarity threshold and $n_1$ is determined by
\begin{equation}
  n_1=\arg \max_i S_k(\mathbf{F}, \mathbf{K}_i)
\end{equation}

\noindent \textbf{Strategy for Similar Examples ($S_v(\mathcal{V}, \mathbf{V}_{n_1}) \ge \delta$).}
For a similar example, the value $\mathbf{V}_{n_1}$ kept unchanged, while the age $A_{n_1} = 0$ and the key $\mathbf{K}_{n_1}$ is updated as follows:

\begin{equation}
  \mathbf{K}_{n_1} = \frac{\mathbf{F} + \mathbf{K}_{n_1}}{\lVert \mathbf{F} + \mathbf{K}_{n_1} \lVert}
\end{equation}
After the memory update, the ages are adjusted as $A_i = A_i + 1$ ($i \neq n_1$).

\noindent \textbf{Strategy for New Examples ($S_v(\mathcal{V}, \mathbf{V}_{n_1}) < \delta$).}
For a new example, the memory writer stores it to the memory network as following

\begin{align}
    \mathbf{K}_{n_o} = \mathbf{F} \\
    \mathbf{V}_{n_o} = \mathcal{V} \\
    A_{n_o} = 0
\end{align}
where $n_o$ is determined by

\begin{equation}
  n_o = \arg \max_i(A_i)
\end{equation}
if there are no empty slots in the memory network.
Otherwise, $n_o$ can be the index of any empty memory slots.
After the memory update, the ages are adjusted as $A_i = A_i + 1$ ($i \neq n_o$).

\subsubsection{Memory Reader}

The memory reader is used for reading the values from the memory network and outputs a value sequence containing 3D volumes that are highly related to the input image features.
For different input image features, there are different numbers of highly similar shapes in the memory.
Therefore, retrieving a fixed number of shapes from the memory network would not be suitable for all inputs and may introduce irrelevant shapes.

To solve this problem, we construct the retrieved value sequence by concatenating all values whose key satisfies $S_k(\mathbf{F}, \mathbf{K}_{n_i}) > \beta $, which can be formulated as

% TODO: Check concatenation
\begin{equation}
  \mathbb{V} = \left[V_{n_i} | S_k(\mathbf{F}, \mathbf{K}_{n_i}) > \beta \right]
  \label{eq:retrieve}
\end{equation}
where $\beta$ is threshold and $[\cdot]$ denotes the concatenation.

\subsection{LSTM Shape Encoder}

The value sequence $\mathbb{V}$ retrieved by the memory reader contains 3D shapes that are similar to the object in the input image.
The value sequence from the memory reader is length-variant and has been ordered by the similarities.
Intuitively, different parts of different shapes in the value sequence may have a different importance in reconstructing the 3D shape from the current image.
To contextually consider and incorporate knowledge useful for current reconstruction from the value sequence into the image feature to supplement the occluded or noisy parts, we leverage LSTM~\cite{lstm} to encode the value sequence $\mathbb{V}$ in a sequential manner.
The LSTM shape encoder takes the length-variant value sequence as input and outputs a fixed-length ``shape prior vector''.
The ``shape prior vector'' is then concatenated with the input image feature to provide extra useful information for the shape decoder.

\subsection{Network Architecture}

\noindent\textbf{Image Encoder.}
The image encoder contains the first three convolutional blocks of ResNet-50~\cite{resnet} to extract a $512 \times 28^2$ feature map from a $224 \times 224 \times 3$ image.
Then the ResNet is followed by three sets of 2D convolutional layers, batch normalization layers and ReLU layers. The kernel sizes of the three convolutional layers are $3^2$, with a padding of 1.
There is a max pooling layer with a kernel size of $2^2$ after the second and third ReLU layers.
The output channels of the three convolutional layers are 512, 256, and 256, respectively.

\noindent\textbf{LSTM Shape Encoder.}
The shape encoder is an LSTM~\cite{lstm} network with 1 hidden layer.
The hidden size is set to 2,048 which indicates that the output shape prior vector is a 2,048 dimensional vector. 

\noindent\textbf{Shape Decoder.}
The decoder contains five 3D transposed convolutional layers. The first four transposed convolutional layers are of kernel sizes $4^3$, with strides of 2 and paddings of 1.
The next transposed convolutional layer has a bank of $1^3$ filter. Each of the first four transposed convolutional layers is followed by a batch normalization layer and a ReLU, and the last transposed convolutional layer is followed by a sigmoid function.
The output channel numbers of the five transposed convolutional layers are 512, 128, 32, 8, and 1, respectively. The final output of decoder is a $32^3$ voxelized shape.

\subsection{Loss Functions}

\noindent \textbf{Voxel Triplet Loss.}
We propose a voxel triplet loss that helps to retrieve precise values from the memory network by guaranteeing that images with similar 3D shapes are closer in the feature space.
In the memory network, $n_p$ and $n_b$ are the memory slots of the positive and negative samples, respectively.
For a positive sample, the similarity between its value $\mathbf{V}_{n_p}$ and the corresponding 3D volume of the input image $\mathcal{V}$ satisfies

\begin{equation}
  S_v(\mathcal{V}, \mathbf{V}_{n_p}) \ge \delta
  \label{eq:similarity}
\end{equation}
Similarly, for a negative sample, the similarity satisfies 

\begin{equation}
  S_v(\mathcal{V}, \mathbf{V}_{n_b}) < \delta
\end{equation}
where $\mathbf{V}_{n_b}$ represents the value of the ``image-voxel'' pair for $n_b$.
Therefore, the voxel triplet loss can be defined as

\begin{equation}
  \ell_t(S_{kb}, S_{kp}, \alpha) = \max
  \left(
  S_{kb} - S_{kp} + \alpha, 0
  \right)
  \label{eq:triplet-loss}
\end{equation}
where $\alpha$ is the margin in the triplet loss~\cite{TripletLoss}.
$S_{kb}$ and $S_{kp}$ are the similarities between the input image features and the keys of the postive/negative sample, 
which are defined as
$S_{kb} = S_k(\mathbf{F},{\mathbf{K}_{n_b})}$ and 
$S_{kp} = S_k(\mathbf{F},{\mathbf{K}_{n_p})}$, respectively.
The proposed voxel triplet loss can minimize the distance among image features with similar 3D volumes and maximize the distance among image features with different 3D volumes. 

\noindent \textbf{Binary Cross Entropy Loss.}
For the reconstruction network, we adopt the Binary Cross Entropy Loss, which is defined as the mean value of the voxel-wise binary cross entropies between the reconstructed object and the ground truth.
More formally, it can be defined as

\begin{equation}
    \ell_r(p, gt) = \frac{1}{r_v^3}\sum_{i=1}^{r_v^3}
    \left[gt_i\log(p_i) + (1 - gt_i)\log(1-p_i)\right]
\end{equation}
where $p$ and $gt$ denote the predicted 3D volume and the corresponding ground truth, respectively. 

The Mem3D is trained end-to-end by the combination of the voxel triplet loss and the reconstruction loss:

\begin{equation}
    \ell_{total} = \ell_t+\ell_r
\end{equation}

\section{Experiments}

\begin{table*}[!th]
\caption{Comparison of single-view 3D object reconstruction on ShapeNet. We report the per-category and overall IoU at $32^3$ resolution. The best results are highlighted in bold.}
\resizebox{\linewidth}{!} {
    \begin{tabular}{ccccccccc}
    \toprule
    Category & 3D-R2N2~\cite{DBLP:conf/eccv/ChoyXGCS16} & OGN~\cite{Tatarchenko_2017_ICCV} & DRC~\cite{DRC}   & Pixel2Mesh~\cite{wang2018pixel2mesh}& IM-Net~\cite{IMnet}& AttSets~\cite{AttSets} & Pix2Vox~\cite{Xie_2019_ICCV} & Mem3D\\
    \midrule
    Airplane & 0.513 & 0.587 & 0.571  &0.508 & 0.702&0.594 & 0.674 & $\textbf{0.767}$ \\
    Bench & 0.421 & 0.481 & 0.453  &0.379  &0.564 &0.552 & 0.608 & $\textbf{0.651}$ \\
    Cabinet & 0.716 & 0.729 & 0.635  & 0.732 &0.680 &0.783 & 0.799 & $\textbf{0.840}$ \\
    Car & 0.798 & 0.828 & 0.755  & 0.670 &0.756 &0.844 & 0.858 & $\textbf{0.877}$ \\
    Chair & 0.466 & 0.483 & 0.469  & 0.484 & 0.644&0.559 & 0.581 & $\textbf{0.712}$ \\
    Display & 0.468 & 0.502 & 0.419 & 0.582 & 0.585& 0.565& 0.548 & $\textbf{0.631}$ \\
    Lamp & 0.381 & 0.398 & 0.415  & 0.399 & 0.433& 0.445& 0.457 & $\textbf{0.535}$ \\
    Speaker & 0.662 & 0.637 & 0.609  & 0.672 & 0.683&0.721 & 0.721 & $\textbf{0.778}$ \\
    Rifle & 0.544 & 0.593 & 0.608 & 0.468 &0.723 & 0.601& 0.617 & $\textbf{0.746}$ \\
    Sofa & 0.628 & 0.646 & 0.606  &0.622  & 0.694& 0.703& 0.725 & $\textbf{0.753}$ \\
    Table & 0.513 & 0.536 & 0.424  & 0.536 &0.621 &0.590 & 0.620 & $\textbf{0.685}$ \\
    Cellphone & 0.661 & 0.702 & 0.413  & 0.762 & 0.762& 0.743& 0.809 & $\textbf{0.823}$ \\
    Watercraft & 0.513 & 0.632 & 0.556  &  0.471&0.607 & 0.601& 0.603 & $\textbf{0.684}$ \\
    \midrule
    overall & 0.560 & 0.596 & 0.545  & 0.552 &0.659 & 0.642& 0.670 & $\textbf{0.729}$ \\
    \bottomrule
\end{tabular}
}
\label{table:shapenet_IOU}
\end{table*}

\begin{table*}
  \caption{Comparison of single-view 3D object reconstruction on ShapeNet. We report the per-category and overall F-Score@1\%. For voxel reconstruction methods, the points are sampled from triangular meshes generated by the marching cube algorithm. The best results are highlighted in bold.}
  \resizebox{\linewidth}{!} {
    \begin{tabular}{ccccccccc}
    \toprule
    Category & 3D-R2N2~\cite{DBLP:conf/eccv/ChoyXGCS16} & OGN~\cite{Tatarchenko_2017_ICCV} & OccNet~\cite{OccNet}   & Pixel2Mesh~\cite{wang2018pixel2mesh}& IM-Net~\cite{IMnet}& AttSets~\cite{AttSets} & Pix2Vox++~\cite{Pix2Vox++} & Mem3D\\
    \midrule
    Airplane  & 0.412 & 0.487 & 0.494  & 0.376 & 0.598 & 0.489 & 0.583 & $\textbf{0.671}$ \\
    Bench     & 0.345 & 0.364 & 0.318  & 0.313 & 0.361 & 0.406 & 0.478 & $\textbf{0.525}$ \\
    Cabinet   & 0.327 & 0.316 & 0.449  & 0.450 & 0.345 & 0.367 & 0.408 & $\textbf{0.517}$ \\
    Car       & 0.481 & 0.514 & 0.315  & 0.486 & 0.304 & 0.497 & 0.564 & $\textbf{0.590}$ \\
    Chair     & 0.238 & 0.226 & 0.365  & 0.386 & 0.442 & 0.334 & 0.309 & $\textbf{0.503}$ \\
    Display   & 0.227 & 0.215 & 0.468  & 0.319 & 0.466 & 0.310 & 0.296 & $\textbf{0.498}$ \\
    Lamp      & 0.267 & 0.249 & 0.361  & 0.219 & 0.371 & 0.315 & 0.315 & $\textbf{0.403}$ \\
    Speaker   & 0.231 & 0.225 & 0.249  & 0.190 & 0.200 & 0.211 & 0.152 & $\textbf{0.262}$ \\
    Rifle     & 0.521 & 0.541 & 0.219  & 0.340 & 0.407 & 0.524 & 0.574 & $\textbf{0.626}$ \\
    Sofa      & 0.274 & 0.290 & 0.324  & 0.343 & 0.354 & 0.334 & 0.377 & $\textbf{0.434}$ \\
    Table     & 0.340 & 0.352 & 0.549  & 0.502 & 0.461 & 0.419 & 0.406 & $\textbf{0.569}$ \\
    Cellphone & 0.504 & 0.528 & 0.273  & 0.485 & 0.423 & 0.469 & 0.633 & $\textbf{0.674}$ \\
    Watercraft& 0.305 & 0.328 & 0.347  & 0.266 & 0.369 & 0.315 & 0.390 & $\textbf{0.461}$ \\
    \midrule
    Overall & 0.351 & 0.368 & 0.393  & 0.398   & 0.405 & 0.395 & 0.436 & $\textbf{0.517}$ \\
    \bottomrule
    \end{tabular}
    
  }
  \label{table:shapenet_fscore}
\end{table*}

\subsection{Datasets}
\noindent\textbf{ShapeNet.}
The ShapeNet dataset~\cite{DBLP:journals/corr/ChangFGHHLSSSSX15} is composed of synthetic images and corresponding 3D volumes.
We use a subset of the ShapeNet dataset consisting of 44K models and 13 major categories following~\cite{DBLP:conf/eccv/ChoyXGCS16}.
Specifically, we use renderings provided by 3D-R2N2 which contains 24 random views of size $137\times137$ for each 3D model.
We also apply random background augmentation~\cite{Pix2Vox++,RenderForCNN} to the image during training.
Note that only the ShapeNet dataset is used for training Mem3D.

\noindent\textbf{Pix3D.} The Pix3D~\cite{DBLP:conf/cvpr/Sun0ZZZXTF18} dataset contains 395 3D models of nine classes.
Each model is associated with a set of real images, capturing the exact object in diverse environments.
The most significant category in this dataset is chairs.
The Pix3D dataset is used only for evaluation.
 
\subsection{Evaluation Metrics}
We apply the intersection over union (IoU) and F-score evaluation metrics widely used by existing works. The IoU is formulated as

\begin{equation}
    \text{IoU} = \frac{\sum_{i,j,k}\mathbb{I}(p(i,j,k)>t)\mathbb{I}(gt(i,j,k))}{\sum_{i,j,k}\mathbb{I}[\mathbb{I}(p(i,j,k)>t)+\mathbb{I}(gt(i,j,k))]}
\end{equation}
where $p(i,j,k)$ and $gt(i,j,k)$ indicate predicted occupancy probability and ground-truth at \emph{(i,j,k)}, respectively.
$\mathbb{I}$ is the indication function which will equal to one when the requirements are satisfied.
The $t$ denotes a threshold, $t=0.3$ in our experiments.
Following Tatarchenko et al. \cite{Oracle}, we also take F-Score as an extra metric to evaluate the performance of 3D reconstruction results, which can be defined as

\begin{equation}
  \textnormal{F-Score}(d) = \frac{2P(d)R(d)}{P(d) + R(d)}
\end{equation}
where $P(d)$ and $R(d)$ denote the precision and recall with a distance threshold $d$, respectively. $P(d)$ and $R(d)$ are computed as

\begin{equation}
  P(d) = \frac{1}{n_{\mathcal{R}}} \sum_{r \in \mathcal{R}} \left[\min_{g \in \mathcal{G}} ||g - r|| < d \right]
\end{equation}

\begin{equation}
  R(d) = \frac{1}{n_{\mathcal{G}}} \sum_{g \in \mathcal{G}} \left[\min_{r \in \mathcal{R}} ||g - r|| < d \right]
\end{equation}
where $\mathcal{R}$ and $\mathcal{G}$ represent the predicted and ground truth point clouds, respectively. $n_\mathcal{R}$ and $n_\mathcal{G}$ are the number of points in $\mathcal{R}$ and $\mathcal{G}$, respectively.
To adapt the F-Score to voxel models, like existing works~\cite{Pix2Vox++}, we apply the marching cube algorithm~\cite{DBLP:conf/siggraph/LorensenC87} to generate the object surface, then 8,192 points are sampled from the surface to compute F-Score between predicted and ground truth voxels.
A higher IoU and F-Score indicates better reconstruction results.

\subsection{Implementation Details}

We used $224\times224$ RGB images as input to train the Mem3D with a batch size of 32.
The whole network is trained end-to-end with the Adam optimizer with a $\beta_1$ of 0.9 and a $\beta_2$ of 0.999.
The initial learning rate is set to 0.001 and decayed by 2 after 150 epochs.
In the memory network, the size is $m = 4000$.
The margin $\alpha$ in Equation \eqref{eq:triplet-loss} is set to $0.1$.
The thresholds $\beta$ and $\delta$ in Equations \eqref{eq:retrieve} and \eqref{eq:similarity} are set to $0.85$ and $0.90$, respectively. The source code will be publicly available.

\subsection{Object Reconstruction on ShapeNet}

We compare the performance with other state-of-the-art methods on the ShapeNet testing set.
Tables~\ref{table:shapenet_IOU} and~\ref{table:shapenet_fscore} show the IoU and F-Score@1\% of all methods, respectively, which indicates that Mem3D outperforms all other competitive methods with a large margin in terms of both IoU and F-Score@1\%.
Our Mem3D benefits from the memory network which explicitly constructs shape priors and applies them according to an object's individual needs to improve reconstruction quality.

\subsection{Object Reconstruction on Pix3D}

Pix3D is a more challenging benchmark which contains diverse real-world images and corresponding shapes.
In Pix3D, the `chair' category contains 3,839 images and the corresponding 3D models, which are the largest category of the dataset.
Due to the complicated environment in images, the objects are frequently occluded by surroundings or themselves.
Therefore, most of the previous works~\cite{Xie_2019_ICCV,DBLP:conf/cvpr/Sun0ZZZXTF18,Wu_2018_ECCV} evaluate their approaches using the hand-selected 2,894 untruncated and unoccluded `chair' images to guarantee their models can capture enough information from the images.
However, although this avoids the occlusion problem to some extent by selecting unoccluded testing samples, the previous reconstruction models still perform imperfectly because of the complicated background.
To show the superior ability to reconstruct objects with the occlusion and background issues, we evaluate Mem3D on the 2,894 chair images with less occlusions but complicated backgrounds and 945 chair images (the complementary set) with heavy occlusions.

Note that the Mem3D is trained on the ShapeNet ``chair'' training set and evaluate on the Pix3D chair set. Since the memory network only writes ``image-voxel'' pairs during training, the memory network only contains shape priors extracted from the ShapeNet dataset.
% Following previous methods~\cite{Pix2Vox++}, we use RenderForCNN~\cite{RenderForCNN} to generate a synthetic dataset by random background augmentation.
% Specifically, we generate 60 images for each 3D CAD model in the ShapeNet dataset.

\subsubsection{Reconstruction with Complicated Backgrounds}
\label{Sec:Reconstruction on Complicated Backgrounds}

\begin{figure}[!t]
    \centering
    \includegraphics[width=1\linewidth]{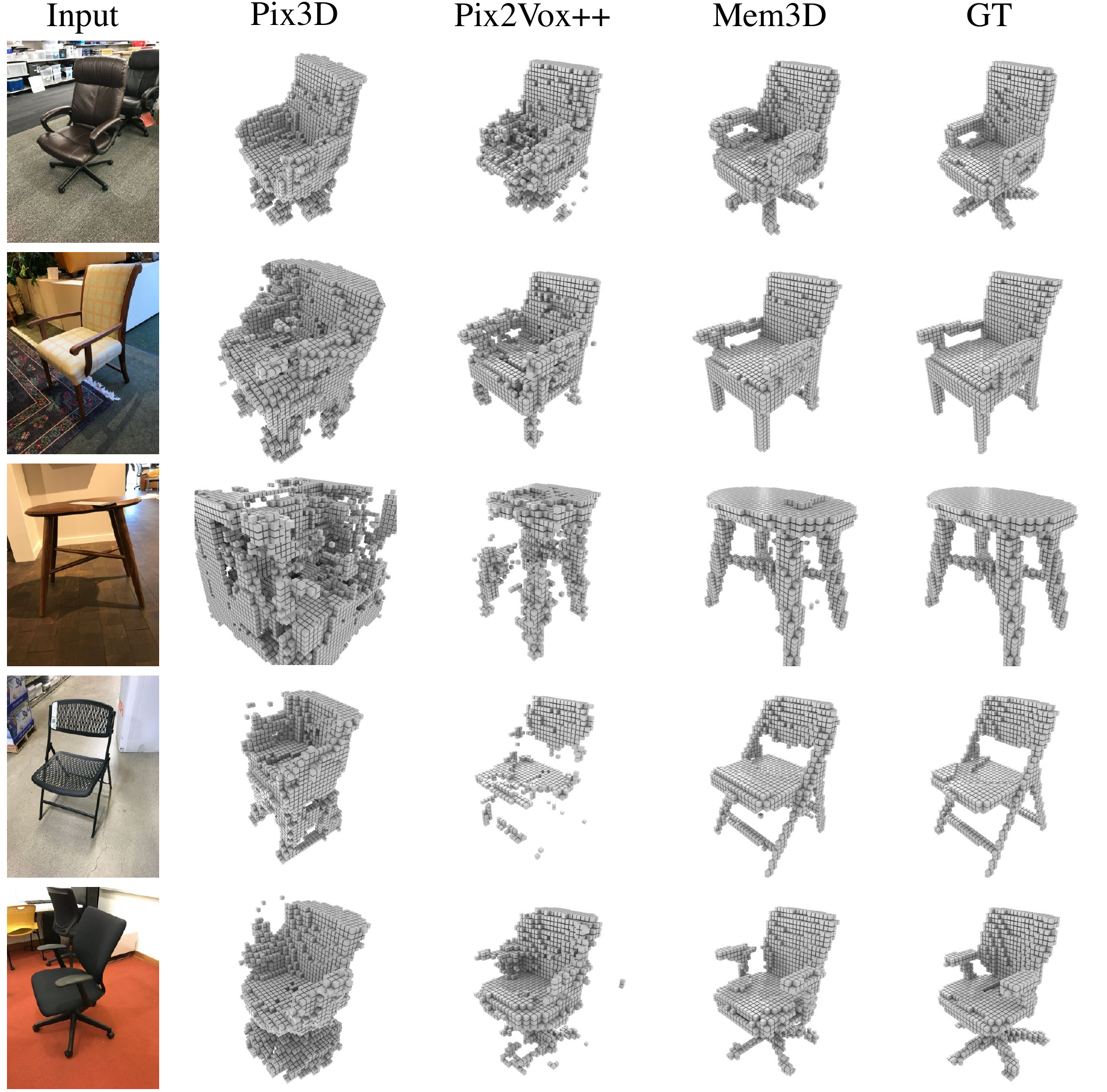}
    \caption{Reconstruction result on 5 of the 2,894 untruncated and unoccluded `chairs' in Pix3D. GT indicates the ground-truth.}
    \label{fig:pix3dv1}
\end{figure}

Table~\ref{pix3d1} shows the evaluation performance of Mem3D and other works on the 2,894 untruncated and unoccluded `chair' images.
Note that these methods use different types of extra information.
For instance, MarrNet~\cite{DBLP:conf/nips/0001WXSFT17}, DRC~\cite{DRC} and ShapeHD~\cite{Wu_2018_ECCV} use extra depth, surface normals and silhouettes information. 
The proposed Mem3D outperforms the state-of-the-art methods by a large margin in terms of both IoU and F-Score@1\%.
The reconstruction results of our Mem3D and previous state-of-the-art works `Pix2Vox++'~\cite{Pix2Vox++} and `Pix3D'~\cite{DBLP:conf/cvpr/Sun0ZZZXTF18} are shown in Figure~\ref{fig:pix3dv1}.
Compared to `Pix2Vox++'~\cite{Pix2Vox++} which employs an encoder-decoder structure, Mem3D can produce more clean and complete reconstruction results.
The reconstructions from Mem3D also provide more details compared to other models.
The memory network in Mem3D can explicitly store and utilize 3D volumes thus providing the reconstruction network with more detailed information about the object and eliminating the background noise.

\begin{table}
\caption{Comparison of single-view 3D object reconstruction on 2,894 untruncated and unoccluded `chair' images in Pix3D.  We report IoU and F-Score@1\%. The best performance is highlighted in bold.}\label{pix3d1}
\vspace{1mm}
\begin{tabularx}{\linewidth}{l|YY}
\toprule
Method  & IoU & F-Score@1\%\\
\midrule
3D-R2N2\cite{DBLP:conf/eccv/ChoyXGCS16} & 0.136 & 0.018 \\
3D-VAE-GAN~\cite{DBLP:conf/nips/0001ZXFT16} & 0.171 & - \\ 
MarrNet~\cite{DBLP:conf/nips/0001WXSFT17} & 0.231 & 0.026\\
DRC~\cite{DRC} & 0.265 & 0.038\\
ShapeHD~\cite{Wu_2018_ECCV} & 0.284 & 0.046\\
DAREC~\cite{Pinheiro_2019_ICCV} & 0.241 & - \\ 
Pix3D~\cite{DBLP:conf/cvpr/Sun0ZZZXTF18} & 0.282 & 0.041 \\
Pix2Vox++~\cite{Pix2Vox++} & 0.292 & 0.068\\
FroDo~\cite{FroDo} & 0.325 & - \\
Mem3D & \textbf{0.387}  & \textbf{0.143}\\
\bottomrule
\end{tabularx}
\vspace{-5mm}
\end{table}

\subsubsection{Reconstruction with Heavy Occlusions}

\begin{figure}[!t]
    \centering
    \includegraphics[width=1\linewidth]{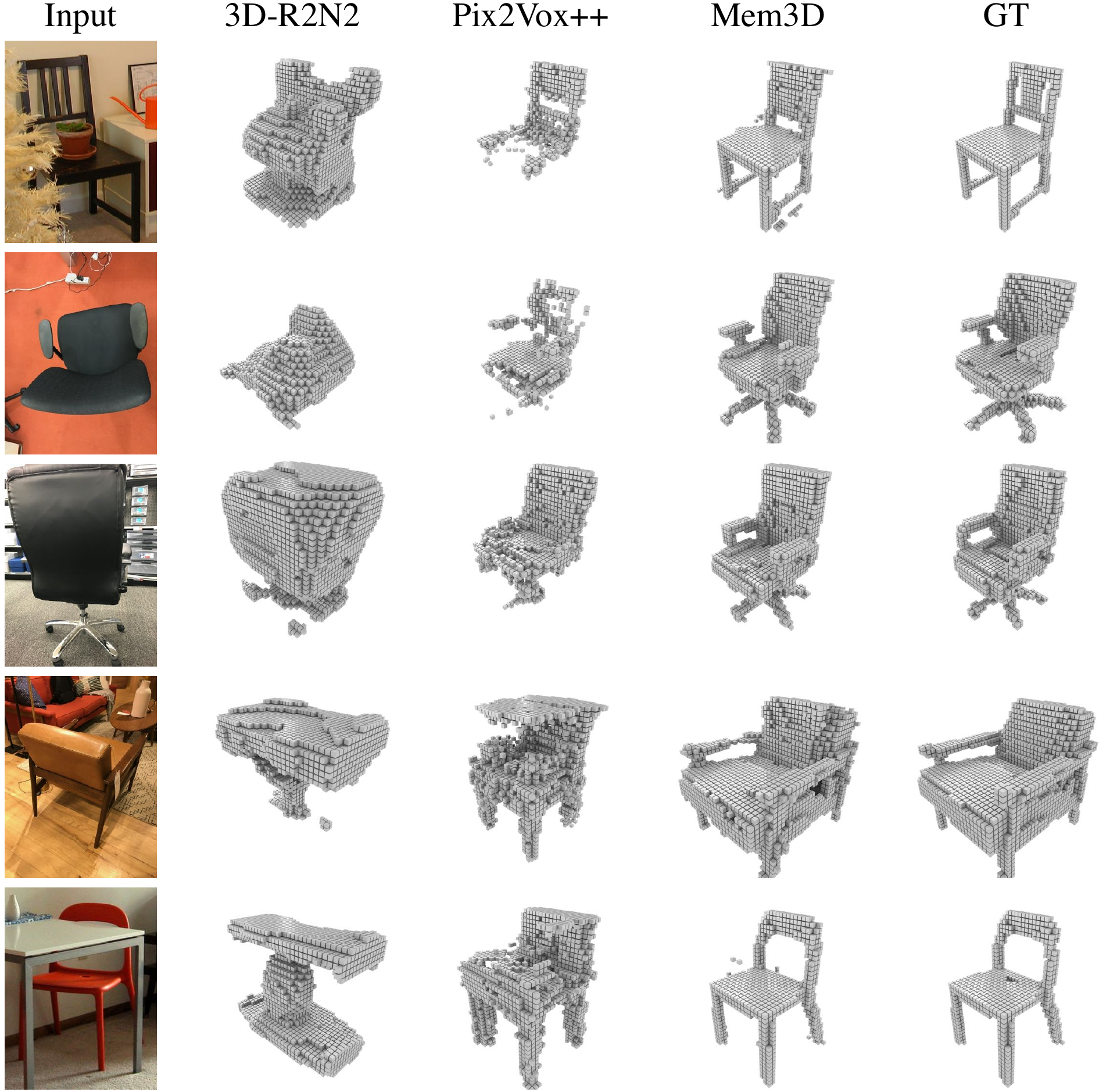}
    \caption{Reconstruction result on 5 of the 945 `chair' images with heavy occlusions in Pix3D. GT indicates the ground-truth.}
    \label{fig:pix3dv2}
\end{figure}

The occlusion issue is another key difficulty for single-view object reconstruction.
Table~\ref{pix3d2} shows the evaluation performance of Mem3D and other works on the 945 chair images with heavy occlusions.
The performance of all other methods drop significantly compared to Section~\ref{Sec:Reconstruction on Complicated Backgrounds}.
While Mem3D shows a favorable ability to handle the extremely cases. Figure~\ref{fig:pix3dv2} shows some reconstruction results of our Mem3D, `Pix2Vox++'~\cite{Pix2Vox++}, and 3D-R2N2~\cite{DBLP:conf/eccv/ChoyXGCS16}.
It can be observed that Pix2Vox++ can reconstruct the perfectly presented parts of object in the image, but failed to reconstruct the occluded parts.
Our Mem3D can provide reasonable reconstruction even for object parts that are hidden in the image.
This is because Mem3D not only captures object information from images, but also obtains complete and clean shape information from the shapes read from memory.
The retrieved shapes can provide detailed and complete shape information for the reconstruction network.

\begin{table}
\caption{Comparison of single-view 3D object reconstruction on 945 `chairs' with heavy occlusions in Pix3D.  We report IoU and F-Score@1\%. The best performance is highlighted in bold.}
\vspace{1mm}
\begin{tabularx}{\linewidth}{l|YY}
\toprule
Method  & IoU & F-Score@1\%\\
\midrule
3D-R2N2\cite{DBLP:conf/eccv/ChoyXGCS16} & 0.055 & 0.011 \\
MarrNet~\cite{DBLP:conf/nips/0001WXSFT17} & 0.138 & 0.019\\
DRC~\cite{DRC} & 0.151 & 0.025\\
ShapeHD~\cite{Wu_2018_ECCV} & 0.183 & 0.037\\
Pix2Vox++~\cite{Pix2Vox++} & 0.215 & 0.041\\
Mem3D & \textbf{0.336} & \textbf{0.105}\\
\bottomrule
\end{tabularx}
\vspace{-5mm}
\label{pix3d2}
\end{table}

\begin{figure*}[!th]
    \centering
    \includegraphics[width=1\linewidth]{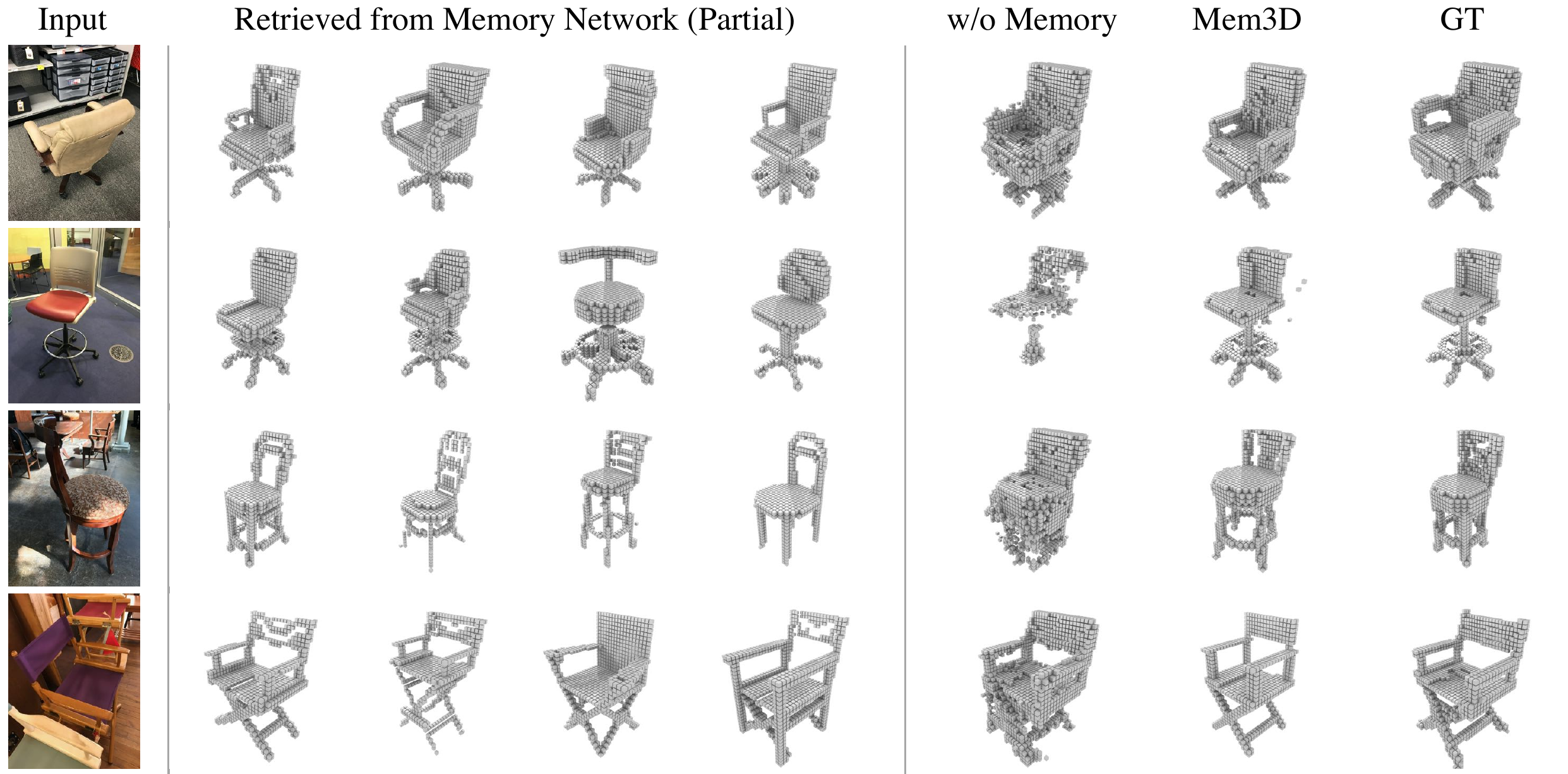}
    \caption{An illustration of the retrieved 3D volumes and the corresponding reconstructions. ``w/o Memory'‘ indicates the reconstruction results are generated without the memory network. We only show the top-4 high-relative 3D volumes retrieved from the memory network. GT indicates the ground-truth.}
    \label{fig:retrieved}
    \vspace{-1mm}
\end{figure*}

\subsection{Ablation Study}

In this section, we evaluate the importance of individual components by ablation studies.

\noindent\textbf{Memory Network.}
To quantitatively evaluate the memory network, we remove the memory network and directly employ the image encoder and decoder as the baseline model.
To further prove the effectiveness of the proposed voxel triplet loss $\ell_t$ in \eqref{eq:triplet-loss}, which pulls image features with similar 3D shapes closer, we remove the voxel triplet loss $\ell_t$ from Mem3D training stage.
The comparison results are shown in Table~\ref{ablation1}, which demonstrates the memory network and the proposed voxel triplet loss contribute significant improvement.
We also show the reconstruction results of the baseline model (without the memory network) and our Mem3D as well as the retrieved shapes in Figure~\ref{fig:retrieved}.
The baseline model which reconstruct the 3D shape of an object from a single image captured in complicated environments are vulnerable to noisy backgrounds and occlusions.
Our proposed Mem3D not only obtains object shapes from images, but also can access complete and clean relevant shapes during reconstruction.
The proposed Mem3D makes it possible to reconstruct the hidden parts of the object in the image and significantly improve the reconstruction quality.

\begin{table} 
\caption{The effect of the memory network and the voxel triplet loss. The best results are highlighted in bold. `m' indicates the memory size and $\ell_t$ indicates the voxel triplet loss in Equation~\eqref{eq:triplet-loss}.}
\vspace{1mm}
  \begin{tabularx}{\linewidth}{l|YY}
  \toprule
  Method  & IoU & F-Score@1\%\\
  \midrule
  w/o memory network & 0.273  & 0.042 \\
  \midrule
  m = 1000  & 0.366 & 0.113\\
  m = 2000  & 0.372 & 0.135\\
  m = 4000 w/o $\ell_t$ &0.359  &0.111 \\
  m = 4000  & \textbf{0.387}  & \textbf{0.143}\\
\bottomrule
\end{tabularx}
\vspace{-4mm}
\label{ablation1}
\end{table}
\begin{table}
  \caption{Different ways of leveraging retrieved shapes. `Top-1' indicates directly treating the first retrieved shape as the reconstruction result. The best results are highlighted in bold.}
  \vspace{1 mm}
  \begin{tabularx}{\linewidth}{l|YY}
    \toprule
    Method              & IoU & F-Score@1\%\\
    \midrule
    Top-1                & 0.287  & 0.051 \\
    Average Fusion       & 0.363 & 0.125 \\
    LSTM Shape Encoder   &  \textbf{0.387}& \textbf{0.143} \\
    \bottomrule
  \end{tabularx}
  \vspace{-5mm}
  \label{ablation2}
\end{table}

\noindent\textbf{LSTM Shape Encoder.}
With the memory network in hand, we have different choices to leverage the retrieved shapes.
For instance, we can directly use the Top-1 retrieved shape as reconstruction result, which is similar to retrieval-based reconstruction~\cite{Oracle}.
We can also use average fusion or the LSTM~\cite{lstm} network to encode useful knowledge from retrieved shapes into a fixed-length vector to condition the decoder.
Table~\ref{ablation2} shows the reconstruction performance when using different ways to leverage the retrieved shapes.
We can also observe the retrieved shapes in Figure~\ref{fig:retrieved}. The Top-1 retrieved shape has similar overall appearance compared to the ground-truth object shape, but the details are very different.
The proposed Mem3D uses the image and the retrieved shapes together to provide high-quality reconstructions, which contains unique details that are distinct from the retrieved 3D volumes.

% \subsection{Discussion}
% To give a detailed analysis of the memory network module, we visualize the top-4 shapes retrieved from the memory and the reconstruction results, as shown in Figure~\ref{fig:retrieved}. To demonstrate the improvement in reconstruction quality brought by the memory module, we train a baseline model which includes an image encoder and decoder but without memory, the reconstructed results of baseline network are shown as `baseline result' in Figure~\ref{fig:retrieved}. The baseline model, as well as the most existing reconstruction models, which obtain object shapes from images captured in complicated environments, are vulnerable to noisy backgrounds and occlusions. The failure cases are shown in Figure~\ref{pix3d1}\ref{fig:pix3dv2} and \ref{fig:retrieved}. Our proposed Mem3D not only obtain object shapes from images, but also can access complete and clean relevant shapes during reconstruction. The introduced memory mechanism makes it possible to reconstruct the hidden parts in the image and significantly improve the reconstruction quality.

\vspace{-3mm}
\section{Conclusion}
\vspace{-2mm}
In this paper, we propose a novel framework for 3D object reconstruction, named Mem3D.
Compared to the existing methods for single-view 3D object reconstruction that directly learn to transform image features into 3D representations, Mem3D constructs shape priors that are helpful to complete the missing image features to recover the 3D shape of an object that is heavy occluded or in a complex environment.
Experimental results demonstrate that Mem3D significantly improves the reconstruction quality and performs favorably against state-of-the-art methods on the ShapeNet and Pix3D datasets.

{\small
\bibliographystyle{ieee_fullname}
\bibliography{egbib}
}

\end{document}